\documentclass[letterpaper, 10 pt, conference]{ieeeconf}  % Comment this line out if you need a4paper
\UseRawInputEncoding

\IEEEoverridecommandlockouts                              % This command is only needed if 
\usepackage{float}
\usepackage{graphicx}
\usepackage{amsmath} % assumes amsmath package installed
\usepackage{amssymb}  % assumes amsmath package installed
\usepackage[normalem]{ulem}
\useunder{\uline}{\ul}{}
\usepackage{booktabs}
\usepackage{footnote}
\usepackage{threeparttable}
\usepackage[colorlinks,linkcolor=red]{hyperref}

\title{\LARGE \bf S-AT GCN: Spatial-Attention Graph Convolution Network based Feature Enhancement for 3D Object Detection
}

\author{Li Wang$^{1}$, Chenfei Wang$^{1}$, Xinyu Zhang$^{1,*}$, Tianwei Lan$^{1}$, Jun Li$^{1}$% <-this % stops a space
	
	\thanks{This work was supported by the National High Technology Research and Development Program of China under Grant No. 2018YFE0204300, and the Beijing Science and Technology Plan Project (Z191100007419008), and the Guoqiang Research Institute Project (2019GQG1010), and the National Natural Science Foundation of China under Grant No. U1964203.}% <-this % stops a space
	\thanks{$^{1}$Li Wang, Chenfei Wang, Xinyu Zhang, Tianwei Lan, Jun Li are with the State Key Laboratory of Automotive Safety and Energy, and the School of Vehicle and Mobility, Tsinghua University, Beijing, 100084 China. (e-mail: wangli\_thu@mail.tsinghua.edu.cn;
	barrycf.wang@gmail.com; 
	xyzhang@tsinghua.edu.cn; 
	jygltw@163.com; 
	lijun19580326@126.com).
Correspondence to:  Xinyu Zhang$<$xyzhang@tsinghua.edu.cn$>$}%
	
}
\begin{document}
\setlength{\abovedisplayskip}{3pt}
\setlength{\belowdisplayskip}{3pt}

	\makeatletter
	\let\@oldmaketitle\@maketitle% Store \@maketitle
	\renewcommand{\@maketitle}{\@oldmaketitle% Update \@maketitle to insert...
		\includegraphics[width=\linewidth]
		{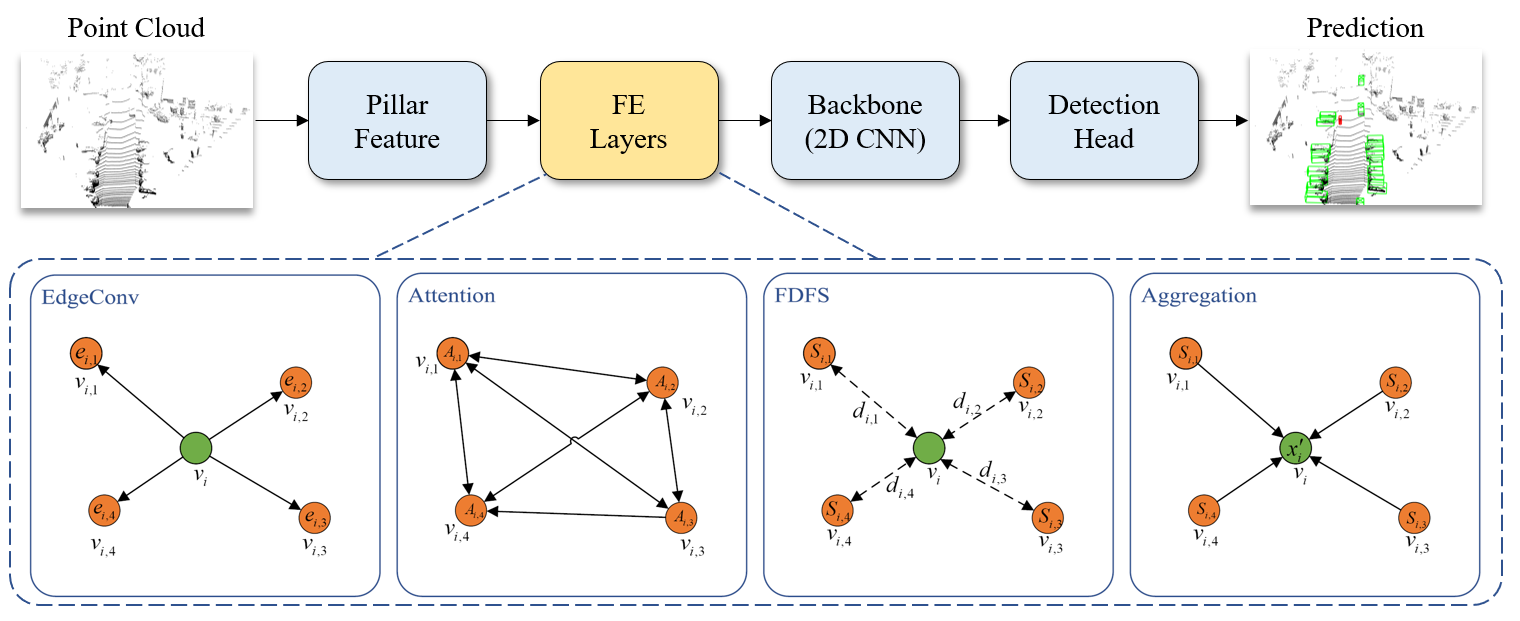}
		Fig. 1. PointPillars with FE layers. FE layers are a plug-in model used to extract detailed geometric information among neighborhoods. In this paper,  FE layers are formed by cascaded Spatial-Attention Graph Convolution (S-AT GCN). S-AT GCN contains 4 steps, including EdgeConv, Attention, FDFS, and Aggregation, corresponding to Vertex Feature Extraction, Self-attention, Far Distance Feature suppression, and Feature Aggregation in Section \ref{subsec_conv}, respectively.
		\setcounter{figure}{1}
		\bigskip}% ... an image
	\makeatother

	\maketitle
	\thispagestyle{empty}
	\pagestyle{empty}

	%%%%%%%%%%%%%%%%%%%%%%%%%%%%%%%%%%%%%%%%%%%%%%%%%%%%%%%%%%%%%%%%%%%%%%%%%%%%%%%%
	\begin{abstract}
		
		3D object detection plays a crucial role in environmental perception for autonomous vehicles, which is the prerequisite of decision and control. This paper analyses partition-based methods' inherent drawbacks. In the partition operation, a single instance such as a pedestrian is sliced into several pieces, which we call it the partition effect. We propose the Spatial-Attention Graph Convolution (S-AT GCN), forming the Feature Enhancement (FE) layers to overcome this drawback. The S-AT GCN utilizes the graph convolution and the spatial attention mechanism to extract local geometrical structure features. This allows the network to have more meaningful features for the foreground. Our experiments on the KITTI 3D object and bird's eye view detection show that S-AT Conv and FE layers are effective, especially for small objects. FE layers boost the pedestrian class performance by 3.62\% and cyclist class by 4.21\% 3D mAP. The time cost of these extra FE layers are limited. PointPillars with FE layers can achieve 48 PFS, satisfying the real-time requirement. The code is available at \url{https://github.com/Link2Link/FE_GCN}.
		
	\end{abstract}

	%%%%%%%%%%%%%%%%%%%%%%%%%%%%%%%%%%%%%%%%%%%%%%%%%%%%%%%%%%%%%%%%%%%%%%%%%%%%%%%%
	\section{Introduction}
	3D Object detection is a core requirement of autonomous vehicles (AVs). To supply the information to the decision and control subsystem, AVs need to scan the surrounding environment. The performance of 3D object detection would directly affect the safety of AVs. To perform the detection task, an autonomous vehicle would typically install several sensors, where the LiDAR is the most significant one. The LiDAR can measure the distance to objects in the surrounding 360-degree view and generate a group of points in three-dimensional space called point cloud. Traditionally, the 3D object detection algorithm is designed as a pipeline involving background subtraction, data clustering and classification \cite{ec1}.
	
	Due to the great success of deep learning in computer vision, there is an increase in the interest to extend this technology to 3D data processing \cite{ec2}. While there are many similarities between 2D object detection in image and 3D object detection in point cloud, they have two key differences: 1) A point cloud is unstructured 3D data, while an image is structured 2D data. As a result, the 2D convolution cannot be easily extended to 3D convolution due to this inherent difference. 2) Points in a point cloud are more spare compared with the pixels in an image. Hence, data in a point cloud are in a sparse representation, while data in an image are dense.
	
	Current 3D object detection algorithms with solely point cloud can be classified into two categories: raw data-based methods and partition-based methods. Raw data-based methods utilize neural networks to process the raw point cloud directly. Partition-based methods apply predefined rules to partition the 3D space (which generates a nominal structure for point cloud) and use a variant of CNN to process the partitioned data.
	
	Raw data-based methods process a point cloud directly. Thus, they do not lose information in theory. Partition-based methods sample a point cloud inside each small zone which we call it cells. If cell features are generated through a statistical method such as mean value, they will suffer from quantization error. If cell feature are learned by the network, even though they do not cause quantization error, the partition-based method will lose some information. Hence, the partition-based methods would inevitably waste some detailed information. We call this phenomenon the partition effect.
	
	In this paper, we analyze the partition effect and point out that the partition operation's drawback is inherent. The partition effect influences the performance of 3D object detection profoundly, especially in small object detection. Based on this observation, we suggest adding an extra layer after partition operation, which we call the Feature Enhancement (FE) layer. We believe that increasing the network capability at very shallow layers to capture geometrical detail information will reduce the partition effect. Based on this intuition, we propose a novel structure called Spatial-Attention Graph Convolution (S-AT GCN), responsible for capturing the geometrical detail in a neighborhood. And the S-AT GCN can be cascaded to form FE layers. To present FE layers' effectiveness, we add it to the baseline model, PointPillars \cite{pointnet}. The evaluation is conducted on the KITTI 3D object detection and birds' eye view detection benchmark. Our results show that FE layers can boost small 3D object detection performance with a tiny cost of time. Our contributions can be summarized as follows:
	\begin{enumerate}
		\item We analyze the partition operation. We show that the partition effect is an inherent drawback that limits small 3D object detection performance in partition-based methods. %The partition effect would drive low-resolution feature maps, which lose the geometrical detail and even the entire geometric structure of an instance. Hence, the traditional convolution cannot effectively extract features from the small object. 
		\item We recommend adding an extra Feature Enhancement (FE) layer behind the partition operation. This extra layer is applied to mitigate the negative impact of the partition effect in small 3D object detection. %The basic idea is that even though the partition effect is inherent, we can generate a more elaborate feature for each voxel or pixel in the pseudo image so that more meaningful detail information can be transferred to the back-end network.
		\item We design a spatial attention mechanism for GCN, which explicitly guarantees the local activation through the Euclidean distance between vertices. %This allows more detailed geometric information been transferred to the back-end network. 
		\item We propose a novel structure called Spatial-Attention Graph Convolution (S-AT GCN) to form the FE layer. S-AT GCN applies graph convolution and spatial attention mechanism to extract detailed geometric information in a local field effectively.
		%\item We experiment on the KITTI 3D object detection benchmark and bird's eye view task. The result shows that compared with the PointPillars baseline, the FE layer brings 3.62\% and 4.21\% (mAP) improvement for the pedestrians and cyclist, respectively.
	\end{enumerate}

	\section{Related Work}
	Most deep learning methods on a point cloud either directly use raw point cloud data \cite{pointnet, pointnet++, ec3, ec4, ec5}, convert the data into 3D grids (voxels) \cite{voxelnet, ec7, ec8, ec9, ec10}, or generate the multi-view representations \cite{ec11, ec12, pointpillars}. Based on the approach to extract features, most methods can be classified into raw data-based and partition-based. The raw data-based methods, just as it literally means, feed the raw point cloud into the network and obtain the result directly. The voxel-based methods and multi-view-based methods can be understood as partition-based methods. The former partition the data in the 3D space to generate the voxel feature, and the latter partition the data in the 2D plane to produce the pixel feature.
	
	\textbf{Raw data-based methods.} PointNet \cite{pointnet} is the first deep learning architecture directly learning the raw point cloud data. It utilizes MLP to learn pointwise features and extracts global features with a max-pooling layer. Since PointNet \cite{pointnet} independently learns features from each point, it cannot capture the local structure. Therefore, Qi et al. \cite{pointnet++} propose a hierarchical network PointNet++ to learn geometric structures from each point's neighborhood. Because of the strong representation ability of PointNet and PointNet++, many 3D object detection algorithms have been developed based on it. \cite{cite1, cite2} take raw point clouds as input and predict bounding boxes for each point. These are two-stage methods. Specifically, set abstraction (SA) layers downsample and extract context features in the first stage. Then, feature propagation (FP) layers are applied for upsampling and broadcasting features to points. Subsequently, the 3D region proposal network (RPN) generates proposals for each point. Based on these proposals, a refinement module is applied to yield the second stage's ultimate prediction. These two-stage methods obtain better performance. However, the inference usually demands a much longer time.
	
	\textbf{Partition-based methods.} Zhou et al. \cite{voxelnet} propose VoxelNet, a voxel-based architecture for learning features from a point cloud. Yan et al. \cite{SECOND} improve VoxelNet by introducing the sparse convolution to obtain the compact representation. \cite{parta2} utilize the multi-task learning to assist the training. PointPillars \cite{pointpillars} introduces a novel feature representation called ``Pillar'' to learn the point distribution in the height-direction. \cite{cite3} attaches the multimodal information into the Pointpilars. Yang et al. \cite{HDNET, PIXOR} project a point cloud to bird's eye view and use a 2D CNN to handle the detection task. All these methods can be interpreted as partition-based methods. First, they partition the point cloud into small cells through a specific predefined rule and apply a statistical approach or learning-based approach to extract features inside these cells. The voxel-based method partition the data into 3D voxels, and the multi-view-based methods partition the data into 2D pixels. Then, deep learning architecture is applied based on these cell features.

	\section{The Drawback of Partition based Method} \label{sec3}
	
	\begin{figure}[htbp]
		\centering
		\includegraphics[width=\linewidth]{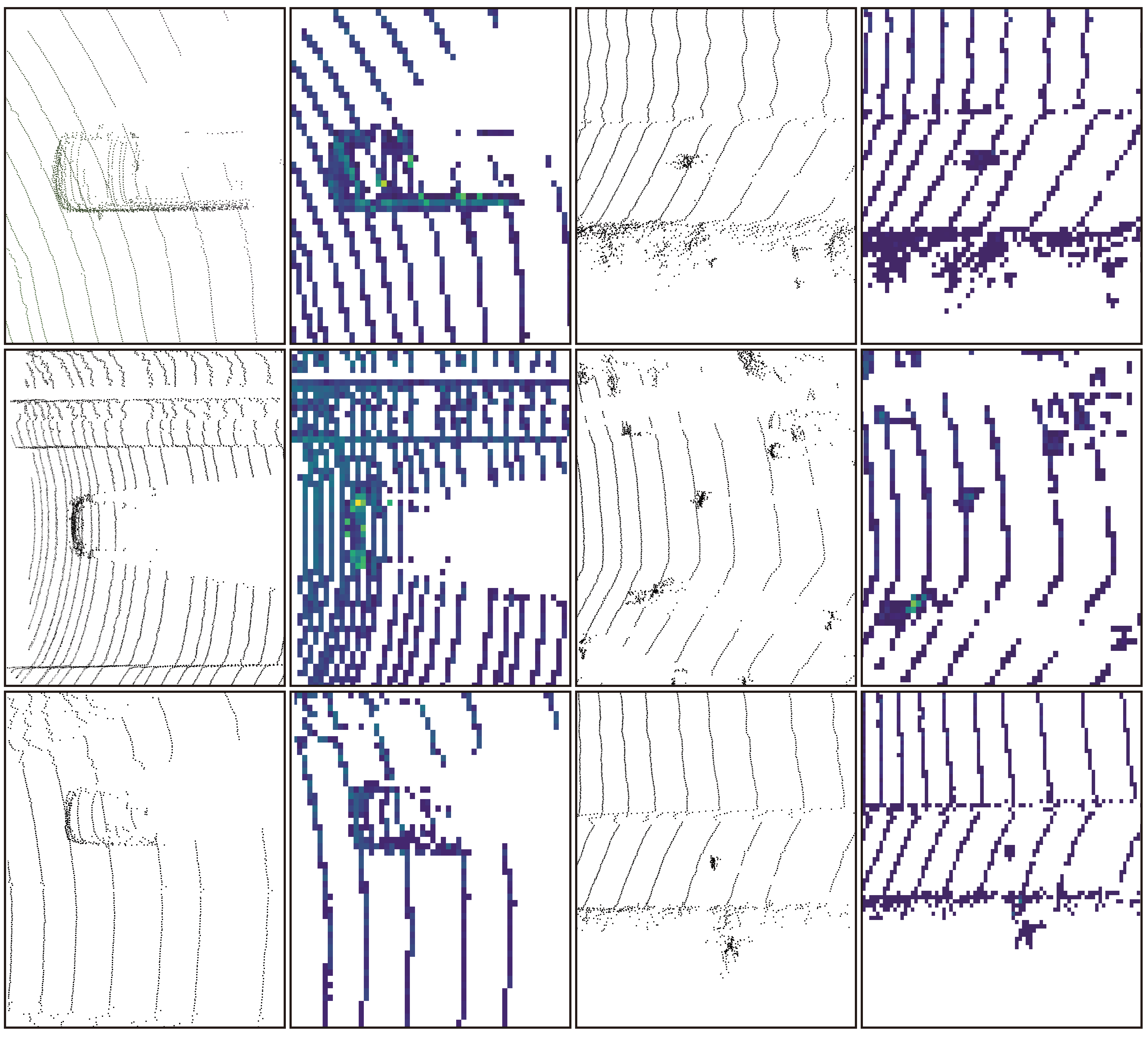}
		\caption{Partition effect for the large object (two columns on left) and small object (two columns on right). After partition and feature extraction, large objects (cars) can maintain the geometrical appearance. But the small objects (pedestrian) almost become mosaic. }
		\label{fig:car_ped}
	\end{figure}
	
	Partition operations have many advantages, including reducing space consumption, reducing computation cost, and transforming unstructured data into structured data. The partition operation, however, has an inherent drawback.
	
	The partition operation slices the space into small pieces. For example, PointPillars\cite{pointpillars} partitions data in a top view and utilizes a simplified PointNet block to produce pillar features. Then, it projects these pillar features onto a bird's eye view to generate a pseudo image. Every pixel in the pseudo image corresponds to a small cell in the raw data space. Inevitably, some instances are sliced into small pieces. This phenomenon is similar to the quantification of voxel-based methods. Since these features are learned from partitioned cells, we call this phenomenon the partition effect. 
	
	For large objects such as cars and trucks, the partition effect may not cause a severe impact in 3D object detection tasks since large objects can maintain the geometrical shape in the pseudo image (left images of Fig. \ref{fig:car_ped}). However, for small objects, the size of cells is comparatively large. Hence, the partition effect becomes more obvious (right image of in Fig. \ref{fig:car_ped}).
	
	Convolution is a spatial-sensitive operation. Since small objects loss their geometrical shapes in the pseudo image, the convolution operation cannot effectively capture small objects' features. Likewise, the same problem also occurs in voxel-based methods, which use 3D convolution. Thus, for all partition-based methods, the partition effect is the main barrier in increasing small objects' detection accuracy.

	The most straightforward way to overcome the drawback of the partition effect is to reduce the cell size. Nevertheless, this will raise space and computational costs. Therefore, we propose a method that improves each cell's feature capturing ability without altering the cell size. Specifically, we propose to add an extra feature extraction layer behind the cell feature extraction to capture more geometrical information. This extra network replaces the role of CNN in shallow layers, which captures the texture information of images. We call this approach Feature Enhancement, and the extra layer is called the Feature Enhancement layer. To achieve the purpose of capturing more geometrical texture information, there are some properties required by the Feature Enhancement layer.
	\begin{enumerate}
		\item Local activation: FE layers are applied to improve the detection ability for detailed information. Hence, the neuron of a FE layer should only activate with data in a local field.
		\item Spatial sensitive: The information of a point cloud is deposited in the local geometrical structure. Therefore, the neuron should be sensitive to the geometrical shape. 
	\end{enumerate}

	\section{Feature Enhancement with S-AT GCN} \label{sec4}
	
	To overcome the shortcoming and keep advantages of partition operation, we propose to add extra FE layers after the partition operation. This section introduces a novel structure called Spatial-Attention Graph Convolution (S-AT GCN), which can be cascaded to form FE layers.

	\subsection{Spatial-Attention GCN} \label{subsec_conv}
	To extract information in the local zone of a point cloud, a straightforward way is to take advantage of the graph convolution network, since a point cloud is an undirected graph naturally. However, we can not represent a point cloud as a fully connected graph in practice because of the huge size of data. So, the edge of the graph needs to be created manually. The most simple strategy of defining the edge is k-nearest neighbors. However, it is not a solid guarantee for local activation. Thus, we introduce the Spatial-Attention mechanism.
	
	A graph $\mathcal{G}$ is a tuple $\mathcal{G}=(\mathcal{V}, \mathcal{E})$ where $\mathcal{V}$ is a set of vertices $v_i \in \mathcal{V}$ and $\mathcal{E} = \{e_{i,j}|v_i,v_j \in \mathcal{V}\}$ represents the connectivity between vertices. All cells and their relationship can be regarded as a graph. Specifically, $v_i=(p_i, x_i)$ is a vertex of the $i_{th}$ cell and it contains cell position $p_i$ and cell feature $x_i$. The set of k-nearest neighbors of the vertex $v_i$ is $\mathcal{N}_i$. The edge of this graph is defined as $\mathcal{E} = \{e_{i,j}|v_i\in\mathcal{V}, v_j\in\mathcal{N}_i \subset \mathcal{V}\} $. 
	
	The following 4 parts (Vertex Feature Extraction, Self-attention with Dimension Reduction, Far Distance Feature Suppression, and Feature Aggregation) are called Spatial-Attention Graph Convolution (S-AT CCN), as shown at the bottom of Fig. 1.
	
	\subsubsection{Vertex Feature Extraction} \label{subsub1}
	DGCNN \cite{dgcnn} introduces an asymmetric edge function that combines the shape structure and neighborhood information. Inspired by EdgeConv, which separates a graph convolution into two parts: edge feature extraction and aggregation, we employ the same asymmetric function to extract features. 
	\begin{equation}
		e^{\prime}_{i,j,m}=ReLU(\theta_m\cdot(x_{i,j}-x_i)+\phi_m \cdot x_i),
	\end{equation}
	where $\theta_m$ and $\phi_m$ are learnable parameters, $m=1,\dots, M$ is the index of $M$-dimensional features. $x_{i,j}$ is the feature of vertex $v_{i,j} \in \mathcal{N}_i$ which is the $j_{th}$ neighbor of $v_i$.
	
	In \cite{dgcnn}, the $e^{\prime}_{i,j,m}$ is understood as the feature of edges. Therefore, it is followed by an aggregation operation in the context of a graph neural network. However, because of $|\mathcal{N}_i|=k$, $e^{\prime}_{i,j,m}$ can also be understood as the feature of $k$ neurons. In this sense, $\mathcal{N}_i$ is actually the reception field of the vertex $v_i$. The following design bases on this observation.
	
	\subsubsection{Self-Attention with Dimension Reduction (ATDR)}
	To enhance the geometric sensitivity of the convolution block, the attention mechanism is applied. \cite{pt,pct} introduce the self-attention mechanism in point cloud processing. In these works, they use a linear projection to generate the query \textbf{Q} and key \textbf{K} \cite{attention} and apply the matrix multiplication as well as activation function to generate the attention map, which can be interpreted as a correlation matrix. Inspired by this idea, we use the linear projection to generate \textbf{Q} and \textbf{K} in one-dimensional feature space.

	\begin{gather}
		Q_{i,j}=\sum_{m=1}^{M}\alpha_{m}e^{\prime }_{i,j,m} \qquad K_{i,j}=\sum_{m=1}^{M}\beta_{m}e^{\prime }_{i,j,m}, \label{f2} \\
		\mathbf{V}_i=\mathbf{Q}_i \times \mathbf{K}_i^T \label{f3},
	\end{gather}
	where $\mathbf{Q}_i =[Q_{i,1}, \dots,Q_{i,k}]^T$, $\mathbf{K}_i =[K_{i,1}, \dots,Q_{i,k}]^T$. $\vec{\alpha}=[\alpha_1,\dots,\alpha_m]^T$ and $\vec{\beta}=[\beta_1,\dots,\beta_m]^T$ are learnable parameters, $\mathbf{Q}_i, \mathbf{K}_i \in \mathbb{R}^{k\times1}$ and $\mathbf{V}_i \in \mathbb{R}^{k\times k}$
	
	\begin{figure}[htbp]
		\centering
		\includegraphics[width=\linewidth]{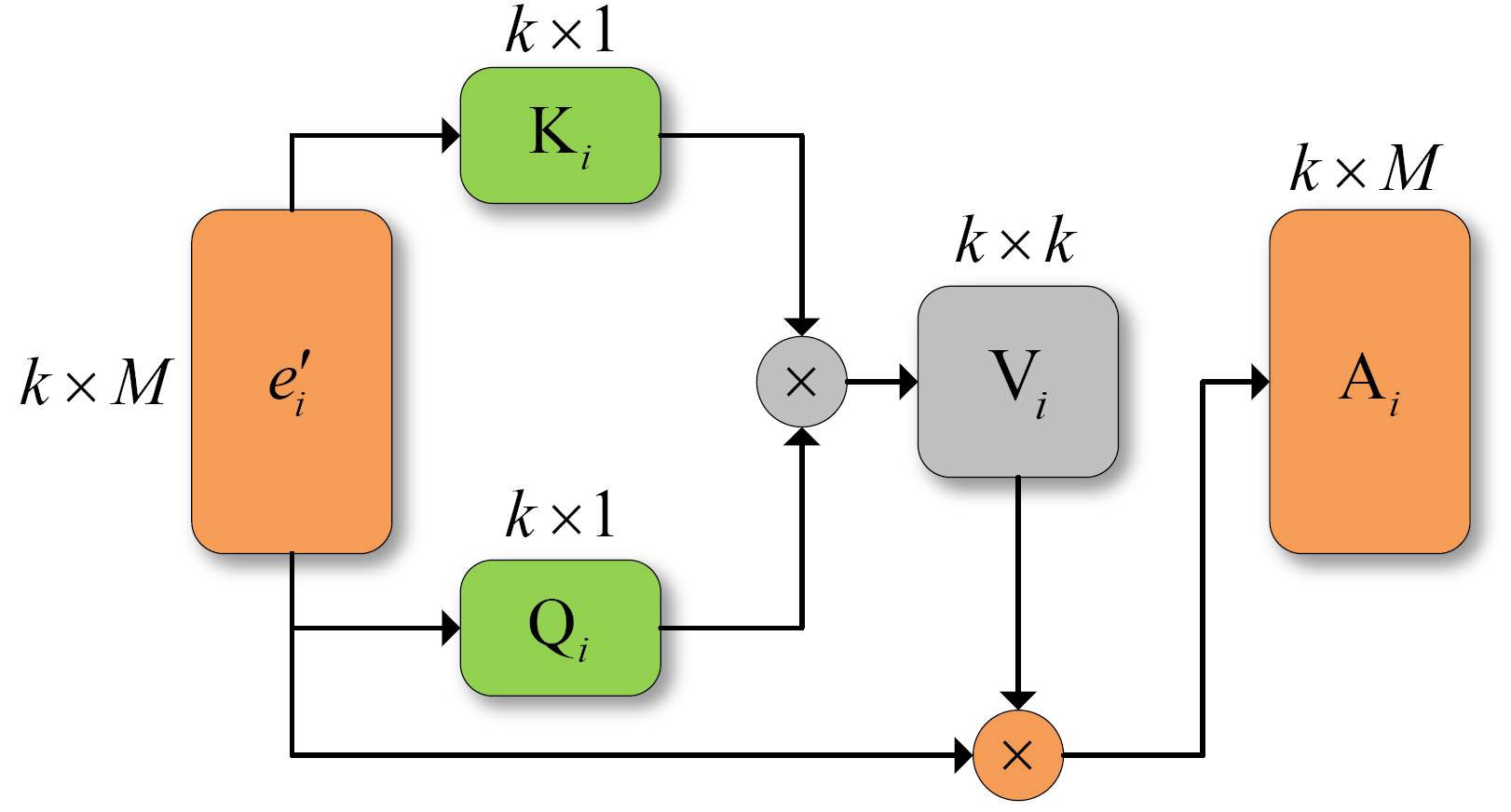}
		\caption{Self-attention with dimension reduction. We use a linear projection to generate the $Q_i$ (query) and $K_i$ (key). $\times$ denotes matrix multiplication. $V_i$ is correlation matrix between query and key. $V_i$ is multiplied with feature $e^\prime_i$ as a part of the self attention mechanism.}
		\label{fig:self-att}
	\end{figure}
	
	\begin{equation}
		\mathbf{A}_i = \mathbf{V}_i \times \mathbf{E}_i^{\prime},
	\end{equation}
	where $\mathbf{E}_i^{\prime}=[e^{\prime}_{i,j,m}] \in \mathbb{R}^{k\times M}$ is the matrix representation of $k$ vertices with $M$-dimensional features. 
	
	The feature $\vec{e}_{i,j} = [e^{\prime }_{i,j,m}] \in \mathbb{R}^M$ is projected to one dimension to generate the $Q_{i,j}$ and $K_{i,j}$. This special design has two benefits:
	
	\begin{itemize}
		\item Physical Intuitive : With the one-dimensional feature for each entry $Q_{i,j}$ and $K_{i,j}$, the matrix $\mathbf{V}_i$ is precisely the correlation matrix between $k$ features. 
		\item Equivalence : Calculating the matrix $\mathbf{V}_i$ from one-dimensional $Q_{i,j}$ and $K_{i,j}$ is equivalent to apply a higher dimensional version in the sense of linear combination. For higher dimensional situation, the scalar multiplication inside the matrix multiplication of (\ref{f3}) becomes an inner product. Hence, the new correlation matrix is $\mathbf{V}_i = \sum_{p=1}^{P} \mathbf{V}_{i,p}$, where $P$ is the dimension of $Q_{i,j}$ and $K_{i,j}$.
	\end{itemize}
	
	\subsubsection{Far Distance Feature Suppression (FDFS)}
	The main information of a point cloud is stored in neighbor data. Thus, we intend the convolution block to activate with a local field. The $k$-nearest neighbors do not strongly guarantee local activation. For example, if cells are very sparse, the $k$-nearest vertices may be far from the vertex $v_i$. Thus, we employ Far Distance Feature Suppression (FDFS) to guarantee the local receptive field explicitly.
	
	\begin{gather}
		d_{i,j}=\left\| p_i-p_j  \right\|_2,  \label{fd} \\
		\sigma_{i,j}=\frac{2}{1+e^{\theta_i \cdot d_{i,j}}}, \label{f5} \\
		\mathbf{S}_{i,j} = \sigma_{i,j} \cdot \mathbf{A}_{i,j}, \label{f6}
	\end{gather}
	where $p_i$ and $p_j$ is the position of vertex $v_i$ and $v_j$. $\theta_i \in \mathbb{R}$ is a learnable parameter. $\mathbf{A}_{i,j} = [A_{i,j, 1}, \dots, A_{i,j, M}]^T \in \mathbb{R}^M$ is the vectorized feature, $A_{i,j,m} = [\mathbf{A}_i]_{j,m}$ is the $(j,m)$ entry of $\mathbf{A}_i$. 
	
	Formula (\ref{fd}) calculates the distance $d_{i,j}$ between vertex $v_i$ and $v_j$. This distance is feeded to sigmoid function (\ref{f5}) to generate the supression parameter $\sigma_{i,j}$. The suppression is applied by a pointwise product (\ref{f6}).
	
	\subsubsection{Feature Aggregation} We apply max-pooling as the aggregation function to compile the features of $\mathcal{N}_i$. 
	\begin{equation}
		x^\prime_{i,m} = \max_{j=1,\dots, k} S_{i,j,m}
	\end{equation}

	\subsection{Feature Enhancement Layer}
	The 4 steps in Section \ref{subsec_conv} form the S-AT GCN, which can be cascaded to constitute the Feature Enhancement (FE) layer. 
	
	\begin{figure}[h]
		\centering
		\includegraphics[scale=0.15]{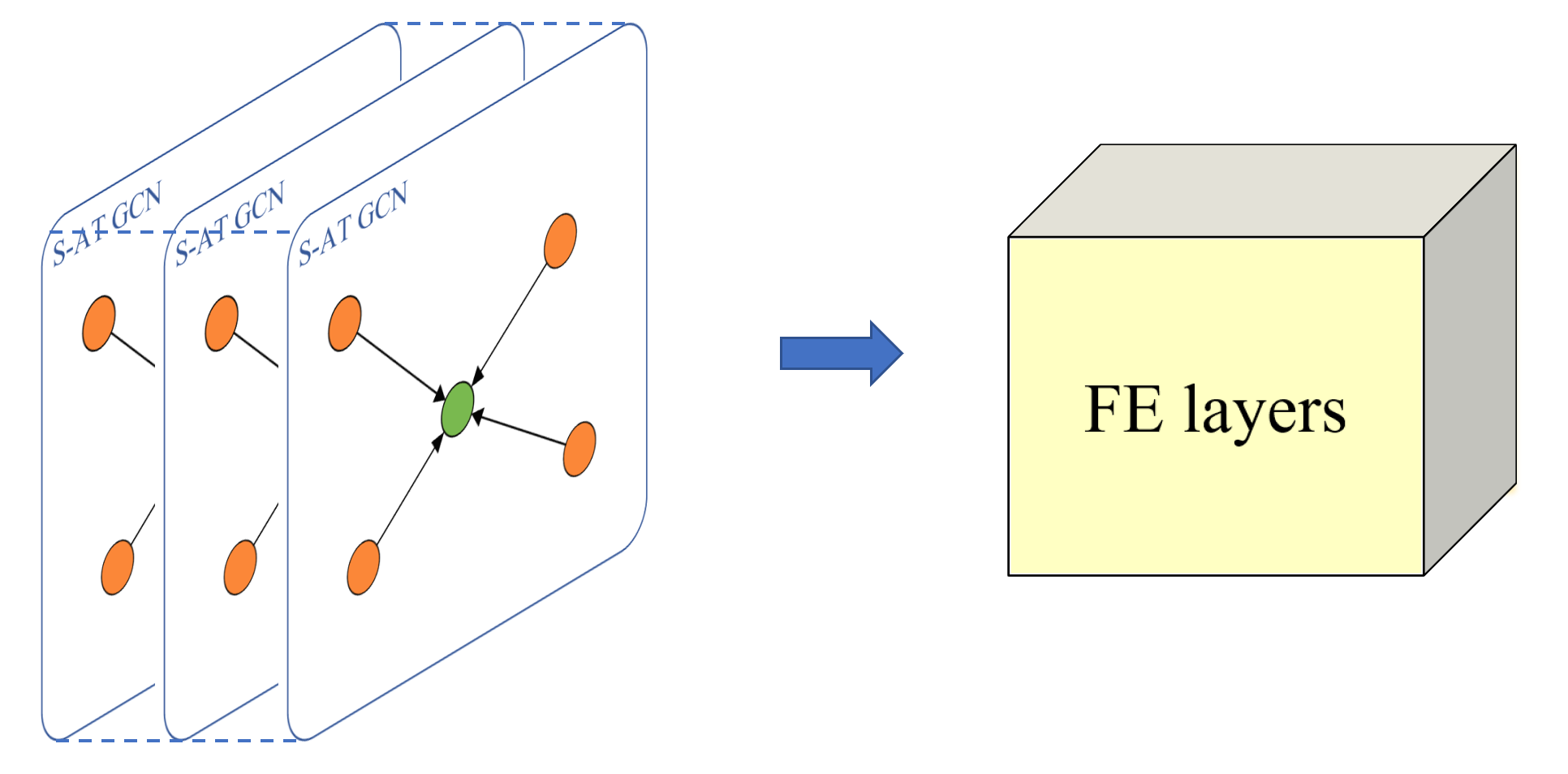}
		\caption{FE layers : Cascaded S-AT GCN to form the Feature Enhancement (FE) layers.}
		\label{fig:fe-layer}
	\end{figure}

	\begin{table*}[htbp]
		\caption{Ablation studies : 3D object detection}
		\centering

    \begin{tabular}{|c|c|ccc|ccc|ccc|ccc|}
    \hline
    Ablation & Models & ATDR & FDFS & Layers & \multicolumn{3}{c|}{Car} & \multicolumn{3}{c|}{Pedestrian} & \multicolumn{3}{c|}{Cyclist} \\
    \multicolumn{1}{|l|}{} & \multicolumn{1}{l|}{} & \multicolumn{1}{l}{} & \multicolumn{1}{l}{} & \multicolumn{1}{l|}{} & \multicolumn{1}{l}{Easy} & \multicolumn{1}{l}{Moderate} & \multicolumn{1}{l|}{Hard} & \multicolumn{1}{l}{Easy} & \multicolumn{1}{l}{Moderate} & \multicolumn{1}{l|}{Hard} & \multicolumn{1}{l}{Easy} & \multicolumn{1}{l}{Moderate} & \multicolumn{1}{l|}{Hard} \\ \hline
    baseline & PointPillars &  &  &  & 87.01 & 77.48 & 75.85 & 55.25 & 50.54 & 46.63 & 78.60 & 62.18 & 62.18 \\ \hline
    FE layer & FE &  &  & 1 & 86.45 & 77.41 & 76.15 & 56.34 & 51.70 & 47.81 & 80.33 & 65.39 & 61.45 \\
     &  &  &  & 3 & 87.11 & 77.00 & 75.57 & 55.01 & 50.79 & 46.87 & 79.13 & 64.49 & 61.82 \\
     &  &  &  & 6 & \textbf{87.56} & 77.56 & \textbf{76.26} & 56.77 & 52.18 & 48.57 & 82.13 & 64.98 & 62.32 \\ \hline
    FDFS & FE:S &  & \checkmark & 1 & 86.98 & \textbf{77.67} & 75.93 & 56.17 & 52.14 & 47.12 & 82.04 & 64.07 & 61.15 \\
     &  &  & \checkmark & 3 & 86.60 & 77.39 & 76.15 & 57.24 & 53.15 & 49.12 & 83.30 & 66.14 & 62.98 \\ \hline
    AT & FE:AT & * &  & 3 & 86.86 & 77.58 & 76.16 & 57.10 & 52.57 & 48.26 & 81.64 & \textbf{67.91} & \textbf{63.89} \\
    ATDR &  & \checkmark &  & 3 & 87.11 & 77.55 & 76.23 & \textbf{58.77} & 53.81 & 49.34 & \textbf{83.40} & 67.27 & 63.85 \\ \hline
    FDFS\&AT & FE:S-AT & * & \checkmark & 3 & 86.03 & 76.95 & 75.52 & 58.52 & \textbf{54.54} & \textbf{50.46} & 82.55 & 67.60 & 62.69 \\
    FDFS\&ATDR &  & \checkmark & \checkmark & 3 & 86.54 & 77.50 & 76.16 & 58.62 & 54.16 & 50.02 & 82.97 & 66.39 & 63.61 \\ \hline
    \end{tabular}

		\label{t1}
		\begin{tablenotes}
			\footnotesize
			\item[1] \checkmark denotes including this term. * denotes self-attention without dimension reduction.
			
		\end{tablenotes}
		
	\end{table*}
	
	\begin{table*}[htbp]
		\caption{Ablation studies: bird's eye view}
		\centering

        \begin{tabular}{|c|c|ccc|ccc|ccc|ccc|}
        \hline
        Ablation & Model & ATDR & FDFS & Layers & \multicolumn{3}{c|}{Car} & \multicolumn{3}{c|}{Pedestrian} & \multicolumn{3}{c|}{Cyclist} \\
         &  &  &  &  & Easy & Moderate & Hard & Easy & Moderate & Hard & Easy & Moderate & Hard \\ \hline
        baseline & PointPillars &  &  &  & 89.54 & 87.28 & 84.90 & 59.98 & 54.33 & 50.02 & 83.01 & 67.65 & 63.38 \\ \hline
        FE layer & FE &  &  & 3 & 89.26 & 87.35 & 85.74 & 62.97 & 57.40 & 53.21 & 84.32 & 70.96 & 66.69 \\
        AT & FE:AT & * &  & 3 & 89.66 & 87.50 & 86.12 & 62.96 & 57.71 & 54.25 & 83.47 & 70.12 & 65.31 \\
        ATDR & FE:AT & \checkmark &  & 3 & 89.54 & 87.56 & \textbf{86.29} & 63.86 & 57.92 & 54.59 & \textbf{85.47} & 71.43 & 68.10 \\
        FDFS\&AT & FE:S-AT & * & \checkmark & 3 & 88.84 & 86.79 & 85.41 & \textbf{64.06} & \textbf{58.93} & 55.24 & 85.19 & 71.06 & 67.10 \\
        FEFS\&ATDR & FE:S-AT & \checkmark & \checkmark & 3 & \textbf{89.70} & \textbf{87.63} & 86.07 & 63.52 & 58.51 & \textbf{55.38} & 84.77 & \textbf{71.80} & \textbf{68.25} \\ \hline
        \end{tabular}

		\label{t2}
		\begin{tablenotes}
			\footnotesize
			\item[1] \checkmark denotes including this term. * denotes self-attention without dimension reduction.
			%\item[2] We compare the results with different setup with the baseline algorithm PointPillars. All results were trained with the same hyper-parameters for 80 epochs on the training set and tested on the validation set of KITTI. 
			
		\end{tablenotes}
		
	\end{table*}

	\textbf{PointPillars with FE Layers.}
	Since FE layers do not change the feature structure, it can be directly plugged into a partition-based model such as PointPillars. FE layers should follow a partition operation.  Thus we add FE layers behind the pillar feature net, which is the partition operation of PointPillars in our context. The FE layer enhances each pillar cell's feature to catch the geometrical structure and information near the neighborhood.

	To implement the S-AT GCN on pillars, we define each pillar as a vertex. The position $p_i$ of vertex $v_i$ is the coordinate of the $i_{th}$ pillar. The feature $x_i$ of vertex $v_i$ is the feature of this pillar. The edge $\mathcal{E}$ is constructed by using $k$-nearest neighbors with $k=9$.
	
	%\subsubsection{Part $A^2$ with FE layer}
	%Part $A^2$ \cite{parta2} uses the mean VFE to quantize the raw point cloud data. In our definition, it is a partition operation. So, for Part $A^2$, the FE layer is added before the 3D sparse convolution.

	\section{Experiment and Results}
	
	\begin{figure*}[ht]
		\centering
		\includegraphics[width=\linewidth]{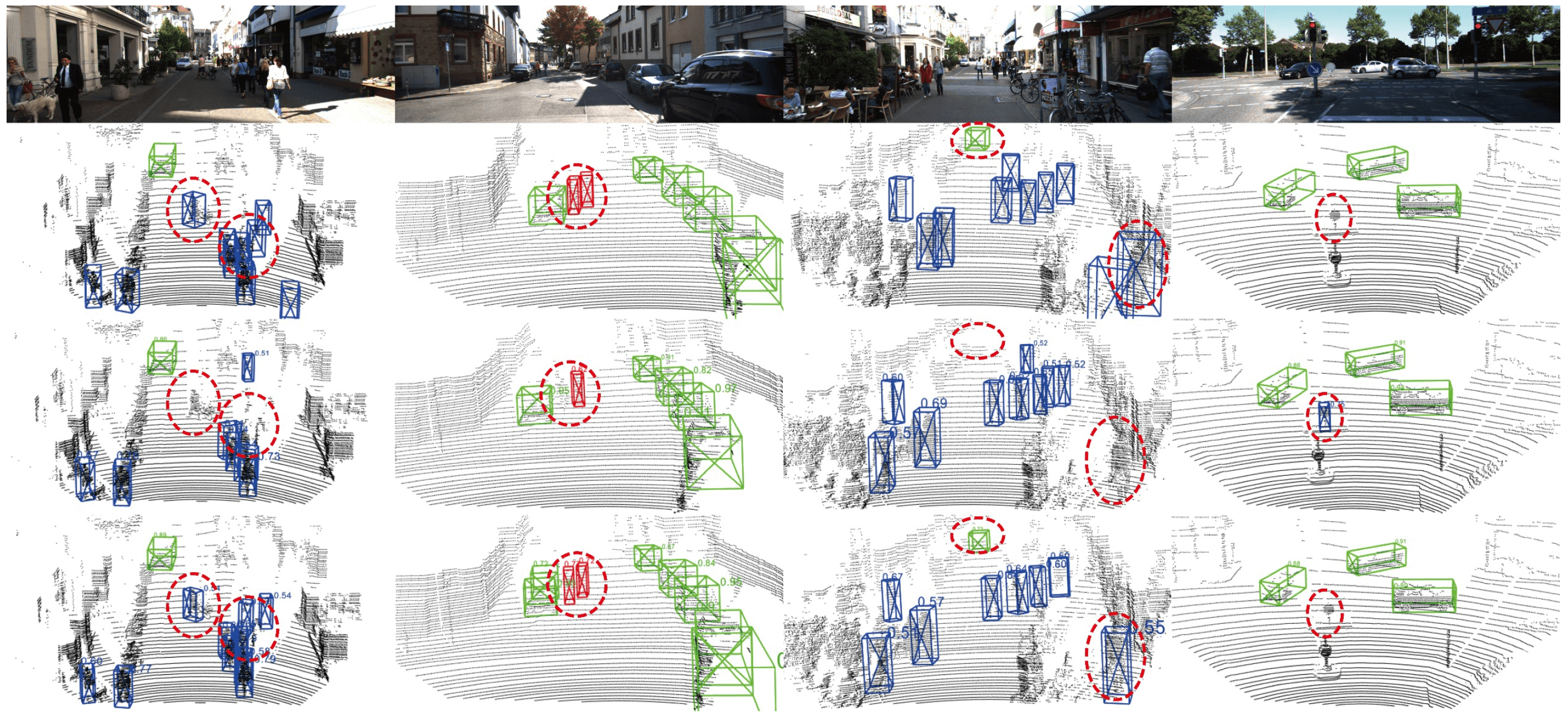}
		\caption{Qualitative results: The first row displays camera images corresponding to each point cloud. The second row gives the ground truth. The third row provides detection results of PointPillars baseline. The last row presents our model's detection results. Solid line boxes with blue, red, and green colors correspond to pedestrian, cyclist, and car, respectively. The red dash line marks the difference between these results.}
		\label{fig:exp}
	\end{figure*}
	
	This section presents our experimental setup and results.
	
	\subsection{Experimental Setup}
	
	We test FE layers on the KITTI 3D Object Detection Benchmark \cite{kitti} and follow \cite{ChenKZBMFU15} to split the provided 7481 training frames into 3712 training samples and 3769 validation samples. For evaluation, we follow the easy, medium, and hard classifications sets proposed by KITTI. 3D detection results are evaluated using the 3D and BEV AP at 0.7 IoU threshold for the car class and 0.5 IoU threshold for the pedestrian and cyclist classes. 
	
	To compare our results with the PointPillars baseline, we follow all network setup of \cite{pointpillars}, including non maximum suppression (NMS) with an overlap threshold of 0.5 IoU, range filter with $[(0, 70.4),(-40, 40),(-3, 1)]$ for car and $[(0, 48), (-20,20), (-2.5, 0.5)]$ for the pedestrian and cyclist. Anchor filter with $(1.6, 3.9, 1.5)$ for car, $(0.6, 0.8, 1.73)$ for pedestrian and $(0.6, 1.76, 1.73)$ for cyclist. The data augmentation is implemented by OpenPCDet \cite{openpcdet2020}.

	\subsection{Effectiveness of FE Layers on PointPillars}
	Our experiments are designed to gradually plug in the S-AT GCN and FE layers to demonstrate the effectiveness of different subassemblies. 
	
	In order to show the effectiveness of the idea of Feature Enhancement, we first add FE layers without the Spatial-Attention mechanism. Then, the ATDR and FDFS are added separately to check the impact of each module. Finally, the full version S-AT GCN is tested.
	
	Results are shown in the TABLE \ref{t1} and TABLE \ref{t2}.  All results are trained with the same hyper-parameters (64 filters, $k=9$) for 80 epochs on the training set and evaluated on the KITTI validation set. 
	
	\textbf{Effect of Feature Enhancement Layers.} Results in Table \ref{t1} show that adding FE layers would improve the detection accuracy of the pedestrian class by $1\%$ and the cyclist class by $2\%-3\%$. At the same time, the detection performance of the car category has not been obviously affected. This demonstrates that FE layers mainly boosts the performance of small 3D object detection. In Table \ref{t2}, the detection accuracy in birds eye's view is raised by $3\%$ and $4\%$ for the pedestrian and cyclist, respectively. The boosted performance in the bird's eye view proves that FE layers can increase the ability of capturing the detailed feature to produce a better pseudo image since the back-end network of PointPillars is CNN.

	\textbf{Effect of Far Distance Feature Suppression.} Results in Table \ref{t1} (FE-FDFS) show that FDFS brings the extra $0.8\%$ performance for pedestrian detection and the extra $0.7\%$ accuracy for cyclist detection. Since the FDFS explicitly restricts each neuron's receptive field, the number of nearest nodes is not significant. Hence, we set $k=9$ in all experiments.

	\textbf{Effect of Self-Attention with Dimension Reduction.} AT and ATDR in Table \ref{t1} denote FE layers with self-attention. The AT indicates self-attention without dimension reduction. The ATDR represents the self-attention mechanism with dimension reduction. With self-attention, FE layers bring extra $1\%$ and $2\%-3\%$ improvement for pedestrians and cyclists, respectively.
	
	\textbf{Layers of FE.} We also study the influence of the number of layers. The number of layers does not show an obvious influence on the performance. Hence, we only show the result of 3 layers for different setups. Since we do not want to lose the real-time requirement, we do not test the performance with deeper FE layers.
	
	\subsection{Qualitative Results}
	Fig. \ref{fig:exp} gives visualization results of the baseline algorithm and our model.  These four rows are image, grand truth, baseline, and our algorithm, respectively. The difference between baseline and our results are marked as dash circles. The first, second, and third columns show the situation that the PointPillars misses some objects. The fourth column gives the situation that PointPillars makes mistakes.
	
	In the scenario of the second column in Fig. \ref{fig:exp}, the PointPillars cannot detect the cyclist in a middle part of the image. The reason is that two cyclists are near each other so that the back-end CNN can not detect these two instances effectively. However, our algorithm can verify these two cyclists successfully. This indicates that, with FE layers, the network is more powerful in capturing the detail. Both the first, second, and third columns in \ref{fig:exp} illustrate a similar situation.
	
	The fourth column in Fig. \ref{fig:exp} presents a case that PointPillars misjudges the traffic sign as a pedestrian. This is a good example to explain the negative influence of the partition effect. The pillar feature net, which is the partition operation in PointPillars, produces a pseudo image in birds' eye view with a low resolution. The traffic sign has a similar appearance to a pedestrian in the pseudo image. Hence, the following CNN cannot distinguish it effectively. However, with FE layers, more detailed geometric information is added to the pseudo image. Even though this instance's shape is not altered in the pseudo image, the contrast is strengthened. Thus, with FE layers, the subsequent CNN is easier to recognize the instance.
	
	The empirical results demonstrate that FE layers boost the performance of PointPillars in terms of increasing the detection (true positive) rate of small objects and sparse objects. Furthermore, it also reduces the wrong detection (false positive) rate.
	
	\begin{figure}[t]
		\centering
		\includegraphics[width=\linewidth]{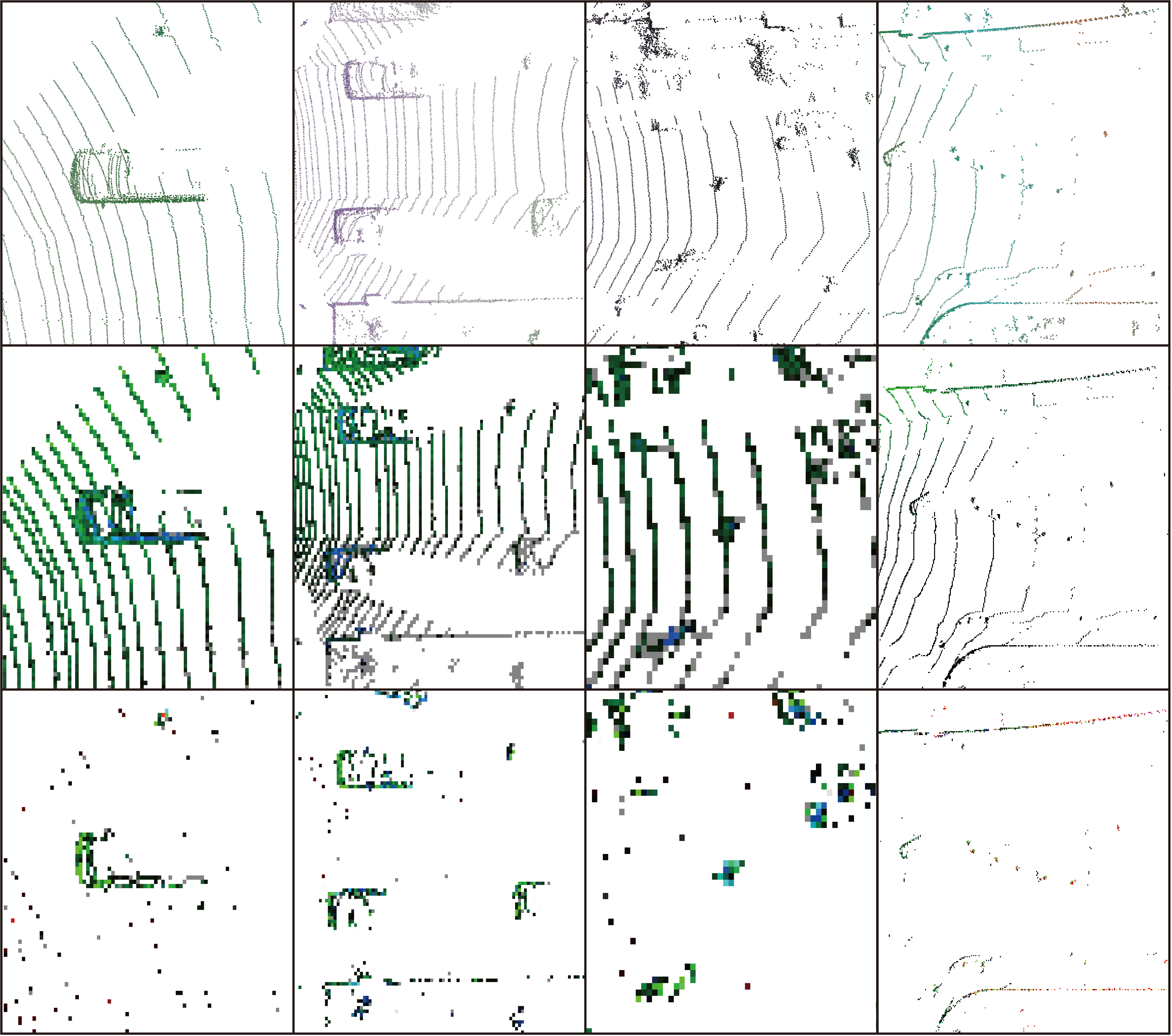}
		\caption{Feature map: Raw point cloud in BEV (top row), feature maps of pillar features (middle row), feature maps of FE layers (bottom row). }
		\label{fig:exp2}
	\end{figure}
	
	Fig. \ref{fig:exp2} gives the visualization of the feature map of PointPillars and PointPillars with FE layers. The pseudo image of PointPillars (middle row) is more like the raw BEV data (top row), compared with the pseudo image of PointPillars with FE layers (bottom row). The feature map with FE layers are insensitive to the variation of the point cloud's density. Moreover, it clearly shows that, with FE layers, the contrast ratio of the feature map is strengthened. At the same time, we can see that most background pixels are empty. This phenomenon hints that FE layers performing as a ``high pass filter''. This observation may help us to understand the role of FE layers in the entire framework. It is responsible for capturing the local geometric information in the very shallow network, which can be explained as detailed texture information.

	%\subsection{Generality of FE layers}
	%To verify the generality of FE layers on small object detection, we also plug it in another baseline algorithm MVX-Net \cite{cite3}. Compared with PointPillars baseline, MVX-Net's voxels are much smaller. Thus, the partition effect is not crucial. However, the table \ref{mvxnet} indicates that FE layers still bring a certain amount of improvement for small objects detection. This demonstrate the generality of FE layers in some extent. FE layers exhibited in the table \ref{mvxnet} are trained with $K=9$. Because of the our hardware's memory limitation, we cannot assess the FE layers in MVX-Net with a larger $K$. The evaluation metric in table is 3D mAP of the moderate set. 
	
    %\begin{table}[h]
	%\centering
	%\caption{Experiment on MVX-Net}		
    %\begin{tabular}{|l|c|c|c|}
    %\hline
    %Models & Car & Ped & Cyc \\ \hline
    %MVX-Net & \textbf{77.12} & 56.65 & 53.18 \\ \hline
    %+FE layers & 76.39 & \textbf{57.02} & \textbf{54.36} \\ \hline
    %\end{tabular}
    %\label{mvxnet}
    %\end{table}
	
	\subsection{Realtime Inference}
	All models are tested on a single card of NVIDIA RTX 3090. The inference time and FPS are shown in the following Table \ref{t3}. With Feature Enhancement layers, the running speed drops to 48 fps. Even though it loses some speed, it can still arrive at the real-time requirements.
	
	\begin{table}[h]
		\centering
		\caption{Running time}
		\begin{tabular}{|c|c|c|}
			\hline
			Models        & ms/image & FPS \\ \hline
			PointPillars & 13.3     & 75  \\ \hline
			FE Layers     & 20.2     & 50  \\ \hline
			ATDR         & 20.8     & 48  \\ \hline
			FDFS\&ATDR   & 20.9     & 48  \\ \hline
		\end{tabular}
		\label{t3}
	\end{table}

	\section{conclusions}
	In this paper, we analyze the partition effect, which limits the performance of partition-based methods in small 3D object detection and propose a structure called Spatial-Attention Graph Convolution (S-AT CCN) to form the Feature Enhancement layer, which is responsible for extracting detailed geometrical information in shallow layers. The proposed method is differentiated from current algorithms by using the GCN and spatial attention mechanism to guarantee the local activation, which is significant in overcoming the partition effect. Experiments on the KITTI dataset show the superiority of our proposed method in small 3D object detection. Finally, the proposed method is shown to run in real-time.

	\addtolength{\textheight}{-0cm}   % This command serves to balance the column lengths
	
	%%%%%%%%%%%%%%%%%%%%%%%%%%%%%%%%%%%%%%%%%%%%%%%%%%%%%%%%%%%%%%%%%%%%%%%%%%%%%%%%

	%%%%%%%%%%%%%%%%%%%%%%%%%%%%%%%%%%%%%%%%%%%%%%%%%%%%%%%%%%%%%%%%%%%%%%%%%%%%%%%%

	%%%%%%%%%%%%%%%%%%%%%%%%%%%%%%%%%%%%%%%%%%%%%%%%%%%%%%%%%%%%%%%%%%%%%%%%%%%%%%%%
	%\section*{APPENDIX}
	
	
	%\section*{ACKNOWLEDGMENT}
	
	%The preferred spelling of the word ÒacknowledgmentÓ in America is without an ÒeÓ after the ÒgÓ. Avoid the stilted expression, ÒOne of us (R. B. G.) thanks . . .Ó  Instead, try ÒR. B. G. thanksÓ. Put sponsor acknowledgments in the unnumbered footnote on the first page.

	%%%%%%%%%%%%%%%%%%%%%%%%%%%%%%%%%%%%%%%%%%%%%%%%%%%%%%%%%%%%%%%%%%%%%%%%%%%%%%%%
	
	%References are important to the reader; therefore, each citation must be complete and correct. If at all possible, references should be commonly available publications.

\bibliographystyle{biblio/ieee}
\bibliography{biblio/IROS}

\begin{thebibliography}{10}\itemsep=-1pt

\bibitem{ChenKZBMFU15}
X.~Chen, K.~Kundu, Y.~Zhu, A.~G. Berneshawi, H.~Ma, S.~Fidler, and R.~Urtasun.
\newblock 3d object proposals for accurate object class detection.
\newblock In C.~Cortes, N.~D. Lawrence, D.~D. Lee, M.~Sugiyama, and R.~Garnett,
  editors, {\em NIPS}, pages 424--432, 2015.

\bibitem{pt}
N.~Engel, V.~Belagiannis, and K.~Dietmayer.
\newblock Point transformer.
\newblock {\em CoRR}, abs/2011.00931, 2020.

\bibitem{kitti}
A.~Geiger, P.~Lenz, and R.~Urtasun.
\newblock Are we ready for autonomous driving? the kitti vision benchmark
  suite.
\newblock In {\em Computer Vision and Pattern Recognition (CVPR), 2012 IEEE
  Conference on}, pages 3354--3361. IEEE, 2012.

\bibitem{pct}
M.-H. Guo, J.-X. Cai, Z.-N. Liu, T.-J. Mu, R.~R. Martin, and S.-M. Hu.
\newblock Pct: Point cloud transformer, 2020.

\bibitem{ec2}
Y.~Guo, H.~Wang, Q.~Hu, H.~Liu, L.~Liu, and M.~Bennamoun.
\newblock Deep learning for 3d point clouds: A survey.
\newblock {\em IEEE Transactions on Pattern Analysis and Machine Intelligence},
  PP:1--1, 06 2020.

\bibitem{ec1}
M.~Himmelsbach, A.~Müller, T.~Luettel, and H.-J. Wuensche.
\newblock Lidar-based 3d object perception.
\newblock 10 2008.

\bibitem{ec3}
M.~Joseph-Rivlin, A.~Zvirin, and R.~Kimmel.
\newblock Mo-net: Flavor the moments in learning to classify shapes.
\newblock {\em CoRR}, abs/1812.07431, 2018.

\bibitem{pointpillars}
A.~H. Lang, S.~Vora, H.~Caesar, L.~Zhou, J.~Yang, and O.~Beijbom.
\newblock Pointpillars: Fast encoders for object detection from point clouds.
\newblock {\em CoRR}, abs/1812.05784, 2018.

\bibitem{pointnet}
C.~R. Qi, H.~Su, K.~Mo, and L.~J. Guibas.
\newblock Pointnet: Deep learning on point sets for 3d classification and
  segmentation, 2017.

\bibitem{pointnet++}
C.~R. Qi, L.~Yi, H.~Su, and L.~J. Guibas.
\newblock Pointnet++: Deep hierarchical feature learning on point sets in a
  metric space.
\newblock In I.~Guyon, U.~von Luxburg, S.~Bengio, H.~M. Wallach, R.~Fergus,
  S.~V.~N. Vishwanathan, and R.~Garnett, editors, {\em NIPS}, pages 5099--5108,
  2017.

\bibitem{ec10}
H.~Rhodin, V.~Constantin, I.~Katircioglu, M.~Salzmann, and P.~Fua.
\newblock Neural scene decomposition for multi-person motion capture, 2019.
\newblock cite arxiv:1903.05684Comment: CVPR 2019.

\bibitem{ec8}
G.~Riegler, A.~O. Ulusoy, and A.~Geiger.
\newblock Octnet: Learning deep 3d representations at high resolutions.
\newblock In {\em CVPR}, pages 6620--6629. IEEE Computer Society, 2017.

\bibitem{parta2}
S.~{Shi}, Z.~{Wang}, J.~{Shi}, X.~{Wang}, and H.~{Li}.
\newblock From points to parts: 3d object detection from point cloud with
  part-aware and part-aggregation network.
\newblock {\em IEEE Transactions on Pattern Analysis and Machine Intelligence},
  pages 1--1, 2020.

\bibitem{cite3}
V.~A. Sindagi, Y.~Zhou, and O.~Tuzel.
\newblock Mvx-net: Multimodal voxelnet for 3d object detection.
\newblock In {\em ICRA}, pages 7276--7282. IEEE, 2019.

\bibitem{openpcdet2020}
O.~D. Team.
\newblock Openpcdet: An open-source toolbox for 3d object detection from point
  clouds.
\newblock \url{https://github.com/open-mmlab/OpenPCDet}, 2020.

\bibitem{attention}
A.~Vaswani, N.~Shazeer, N.~Parmar, J.~Uszkoreit, L.~Jones, A.~N. Gomez,
  Å.~Kaiser, and I.~Polosukhin.
\newblock Attention is all you need.
\newblock In I.~Guyon, U.~V. Luxburg, S.~Bengio, H.~Wallach, R.~Fergus,
  S.~Vishwanathan, and R.~Garnett, editors, {\em Advances in Neural Information
  Processing Systems 30}, page 5998–6008. Curran Associates, Inc., 2017.

\bibitem{ec9}
P.-S. Wang, Y.~Liu, Y.-X. Guo, C.-Y. Sun, and X.~Tong.
\newblock O-cnn: octree-based convolutional neural networks for 3d shape
  analysis.
\newblock {\em ACM Trans. Graph.}, 36(4):72:1--72:11, 2017.

\bibitem{dgcnn}
Y.~Wang, Y.~Sun, Z.~Liu, S.~E. Sarma, M.~M. Bronstein, and J.~M. Solomon.
\newblock Dynamic graph cnn for learning on point clouds.
\newblock {\em ACM Trans. Graph.}, 38(5):146:1--146:12, 2019.

\bibitem{ec7}
Z.~Wu, S.~Song, A.~Khosla, F.~Yu, L.~Zhang, X.~Tang, and J.~Xiao.
\newblock 3d shapenets: A deep representation for volumetric shapes.
\newblock In {\em CVPR}, pages 1912--1920. IEEE Computer Society, 2015.

\bibitem{SECOND}
Y.~Yan, Y.~Mao, and B.~Li.
\newblock Second: Sparsely embedded convolutional detection.
\newblock {\em Sensors}, 18(10):3337, 2018.

\bibitem{HDNET}
B.~Yang, M.~Liang, and R.~Urtasun.
\newblock Hdnet: Exploiting hd maps for 3d object detection.
\newblock In {\em CoRL}, volume~87 of {\em Proceedings of Machine Learning
  Research}, pages 146--155. PMLR, 2018.

\bibitem{PIXOR}
B.~Yang, W.~Luo, and R.~Urtasun.
\newblock Pixor: Real-time 3d object detection from point clouds.
\newblock In {\em CVPR}, pages 7652--7660. IEEE Computer Society, 2018.

\bibitem{ec4}
J.~Yang, Q.~Zhang, B.~Ni, L.~Li, J.~Liu, M.~Zhou, and Q.~Tian.
\newblock Modeling point clouds with self-attention and gumbel subset sampling.
\newblock In {\em CVPR}, pages 3323--3332. Computer Vision Foundation / IEEE,
  2019.

\bibitem{cite2}
Z.~Yang, Y.~Sun, S.~Liu, X.~Shen, and J.~Jia.
\newblock Ipod: Intensive point-based object detector for point cloud.
\newblock {\em CoRR}, abs/1812.05276, 2018.

\bibitem{cite1}
Z.~Yang, Y.~Sun, S.~Liu, X.~Shen, and J.~Jia.
\newblock Std: Sparse-to-dense 3d object detector for point cloud.
\newblock In {\em ICCV}, pages 1951--1960. IEEE, 2019.

\bibitem{ec12}
T.~Yu, J.~Meng, and J.~Yuan.
\newblock Multi-view harmonized bilinear network for 3d object recognition.
\newblock In {\em CVPR}, pages 186--194. IEEE Computer Society, 2018.

\bibitem{ec11}
Y.~Zhang, M.~J. Er, R.~Zhao, and M.~Pratama.
\newblock Multiview convolutional neural networks for multidocument extractive
  summarization.
\newblock {\em IEEE Trans. Cybern.}, 47(10):3230--3242, 2017.

\bibitem{ec5}
H.~Zhao, L.~Jiang, C.-W. Fu, and J.~Jia.
\newblock Pointweb: Enhancing local neighborhood features for point cloud
  processing.
\newblock In {\em CVPR}, pages 5565--5573. Computer Vision Foundation / IEEE,
  2019.

\bibitem{voxelnet}
Y.~Zhou and O.~Tuzel.
\newblock Voxelnet: End-to-end learning for point cloud based 3d object
  detection.
\newblock In {\em CVPR}, pages 4490--4499. IEEE Computer Society, 2018.

\end{thebibliography}

\end{document}